\documentclass[lettersize,journal]{ieeeconf}
\IEEEoverridecommandlockouts
\usepackage{amsmath,amsfonts}
\usepackage{array}
\usepackage[caption=false,font=normalsize,labelfont=sf,textfont=sf]{subfig}
\usepackage{textcomp}
\usepackage{stfloats}
\usepackage{url}
\usepackage{verbatim}
\usepackage{graphicx}
\usepackage{cite}
\usepackage{times}
\usepackage{epsfig}
\usepackage{float}
\usepackage{wrapfig}
\usepackage{algorithm,algorithmicx,algpseudocode}
\usepackage{bm,xspace}
\usepackage{comment}
\usepackage{multirow}
\usepackage{balance}
\usepackage{booktabs}
\usepackage{etoolbox,siunitx}
\usepackage{calc}
\usepackage{pifont,hologo}
\usepackage{nicefrac}
\usepackage{makecell}
\usepackage{adjustbox}

\usepackage{color}

\makeatletter
\let\NAT@parse\undefined
\makeatother
\usepackage[colorlinks=black,citecolor=green]{hyperref}

\begin{document}

\title{City-scale Continual Neural Semantic Mapping with Three-layer Sampling and Panoptic Representation}

\author{Yongliang Shi$^{1*}$, Runyi Yang$^{1,2*}$,  Pengfei Li$^{1}$, Zirui Wu$^{1,2}$, Hao Zhao$^{1}$, Guyue Zhou$^{1\dag}$ 
\thanks{*Equal contribution, \dag  Corresponding author}
\thanks{$^{1}$Institute for AI Industry Research (AIR), Tsinghua University, China,
    \{shiyongliang, lipengfei, zhaohao, zhouguyue\} @air.tsinghua.edu.cn.}%
\thanks{$^{2}$Beijing Institute of Technology, China,  \{wuzirui, runyi.yang\} @bit.edu.cn.}%
\thanks{This work was sponsored by Tsinghua-Toyota Joint Research Fund.}
}


\makeatletter
\let\NAT@parse\undefined
\makeatother
\makeatletter
\g@addto@macro\@maketitle
{
  \begin{figure}[H]
  \setlength{\linewidth}{\textwidth}
  \setlength{\hsize}{\textwidth}
  \setcounter{figure}{0}  
  \centering
  \begin{tabular}{@{}c@{\hspace{1mm}}c@{}}
 	\includegraphics[width=0.95\textwidth]{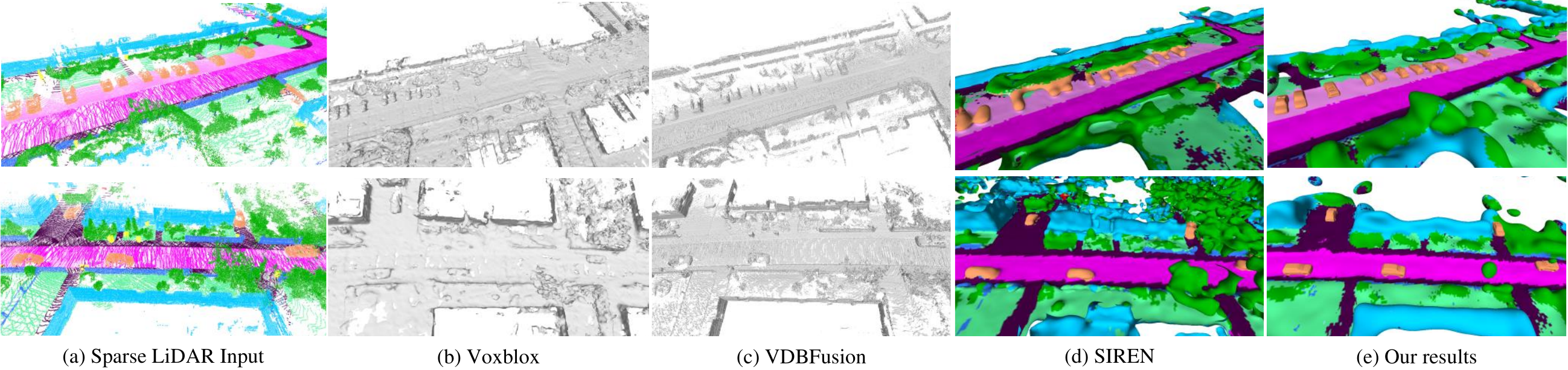}
  \end{tabular}
  \vspace{-0.5mm}
  \caption{(a) The input is sparsity-variant point cloud of road scenes captured by LiDAR. 
  Explicit fitting results by (b) Voxblox\cite{oleynikova2017voxblox} and (c)  VDBFusion\cite{vizzo2022sensors}. (d) The implicit fitting results of SIREN\cite{sitzmann2020implicit} with our semantic prediction model. (e) Our results.  
  }
  \label{fig:teaser}
  \vspace{-10mm}
  \end{figure}
}
\makeatother
\maketitle

\begin{abstract}
Neural implicit representations are drawing a lot of attention from the robotics community recently, as they are expressive, continuous and compact. However, city-scale continual implicit dense mapping based on sparse LiDAR input is still an under-explored challenge. To this end, we successfully build a city-scale continual neural mapping system with a panoptic representation that consists of environment-level and instance-level modelling. Given a stream of sparse LiDAR point cloud, it maintains a dynamic generative model that maps 3D coordinates to signed distance field (SDF) values. To address the difficulty of representing geometric information at different levels in city-scale space, we propose a tailored three-layer sampling strategy to dynamically sample the global, local and near-surface domains. Meanwhile, to realize high fidelity mapping of instance under incomplete observation, category-specific prior is introduced to better model the geometric details. We evaluate on the public SemanticKITTI\cite{semantickitti} dataset and demonstrate the significance of the newly proposed three-layer sampling strategy and panoptic representation, using both quantitative and qualitative results. Codes and model will be publicly available.
\end{abstract}
\section{Introduction}
Mapping is widely recognized as a fundamental environment sensing capability of intelligent robots, e.g., city-scale 3D maps are critical to the localization and planning of autonomous vehicles. A good mapping method should have \textbf{small memory footprint} and \textbf{rich map elements} (e.g., geometry and semantics) while allowing \textbf{continual updating}. In this paper, we propose a city-scale mapping system that meets these three requirements simultaneously, while addressing several non-trivial technical issues.

With these three requirements in mind, the first question to ask is: which representation should we use? People have developed various explicit scene representations for mapping, such as point clouds \cite{rozenberszki2020lol}(Fig.\ref{fig:teaser}(a)), voxel grids\cite{wu2018learning}, octrees\cite{hornung2013octomap}, and surfel clouds \cite{behley2018efficient}. Being explicit means that, although these representations differ in details, they store the coordinates (and other properties) of 3D map points explicitly. As such, they boil down to some forms of discrete approximation of the underlying 3D map, and their memory footprint inevitably grows with the number of 3D map points. 



Contrary to explicit representations, implicitly defined, continuous, differentiable shape representations parameterized by neural network have emerged as a powerful paradigm for surface modeling\cite{mildenhall2020nerf}\cite{takikawa2021neural}\cite{zhong2023icra}. These methods easily deal with a wide variety of surface topologies with arbitrary resolution, enabling downstream tasks ranging from robotic perception\cite{hoeller2022neural} and 3D reconstruction to navigation\cite{adamkiewicz2022vision}. Recently, research about RGBD-based continual implicit mappings has made significant progress\cite{ortiz2022isdf}\cite{yan2021continual}, but they are all used for indoor scene reconstruction. Utilizing implicit representation for dense reconstruction in urban scenes from sparse LiDAR data still suffers from limitations: 
 \textbf{1) Scale variation:} Exclusively relying on a uniform global sampling approach is inadequate in fulfilling the demands for reconstructing details\cite{takikawa2021neural}, especially in urban scenes where the scale varies greatly in different directions, resulting in the difficulty of capturing sufficient local detail and making continual update forgetting surface geometry information (Fig.\ref{fig:teaser}(d)). \textbf{2)} The \textbf{sparsity and incompleteness} of instance data impairs the effectiveness of reconstruction outcomes, particularly when the instance data is severely deficient due to either scan blind spot or occlusion, rendering complete instance reconstruction an arduous task (Fig.\ref{fig:teaser}(b) and Fig.\ref{fig:teaser}(c)).

In this paper, we propose a continual mapping system with panoptic representation under city-scale scene with LiDAR input (Fig.\ref{fig:teaser}(e)). In contrast to implicit simultaneous localization and mapping (SLAM) systems, which focus on updating pose estimation and scene reconstruction, our proposed method prioritizes continual neural semantic mapping. To overcome the memory scalability challenges inherent in traditional explicit reconstruction methods, we employ neural implicit representation. Specifically, we continually update a neural network from sequential LiDAR data, which maps the scene coordinates to signed distance field (SDF) values. This approach allows us to accurately represent the scene while efficiently managing memory usage. 
Furthermore, we jointly train a parallel neural network for semantic segmentation with the geometry model to better support downstream tasks. To mitigate forgetting of surface geometry by neural networks during continual updates, we propose a three-layer sampling method that covers global, local, and near-surface scales. Additionally, we introduce a category-shared prior to optimize a latent code for dense reconstruction, counteracting poor reconstruction quality caused by sparse and incomplete data of instances in the scene.

To summarize, our contributions are as follows:
\begin{itemize}
\item[$\bullet$] A city-scale continual learning system within a panoptic representation consists of scene and instance representation is developed.
\item[$\bullet$] Three-layer sampling method is proposed to facilitate the surface fitting of scene during continual update.
\item[$\bullet$]An instance representation with category-specific prior is put forward to complete dense reconstruction of incomplete and sparse instances. 
\item[$\bullet$] On the city-scale dataset SemanticKITTI\cite{semantickitti}, better performance are demonstrated when using our method.
\end{itemize}
\section{Related Work}
Implicit neural representations are about to parameterize a continuous differentiable signal with a neural network, providing a possibly more compact representation of scenes and instances. There has been much promising recent work on using neural implicit representations. 

\textbf{Scene Representation} 
The Scene Representation Network (SRN)\cite{sitzmann2019scene} was one of the first methods to use a multi-layer perceptron (MLP) as the neural representation of a learned scene given a collection of images and associated poses.
Occupancy Networks \cite{peng2020convolutional}\cite{mescheder2019occupancy} learned an implicit 3D occupancy function for shapes or large-scale scenes gave 3D supervision. SDF has some useful properties, for instance, unlike explicit representations, they allow for changes in surface topology, and they can be updated very easily. A variety of SDF-based implicit neural representations are proposed to  solve boundary value problems. However, ReLU-based MLP \cite{rahaman2019spectral}\cite{xu2019training} are incapable of reconstructing
high-frequency details of surfaces because they are piecewise linear and their derivative is zero. Sitzmann leverages MLPs with periodic activation functions for implicit neural representations\cite{sitzmann2020implicit}. Recently, having adopted a sparse octree-based feature volume to represent surfaces,  SHINE-Mapping\cite{zhong2023icra} enables large-scale incremental
3D mapping and allows for accurate and complete reconstructions. What's more, NeRF\cite{mildenhall2020nerf} has drawn wide attention across scene representation \cite{turki2022mega}\cite{rematas2022urban} due to its simplicity and extraordinary performance.

\textbf{Instance Representation} The aforementioned works mostly focus on learning representations for whole scene within a few categories, and they have not been studied details of instances in large scenes. Jiang\cite{jiang2020local} learn to encode/decode geometric parts of objects at a part scale by training an implicit function auto-encoder, and optimize  Latent Implicit Grid representation that matches a partial  scene observation.  Yang\cite{yang2021learning} used a shared MLP with instance-specific latent codes to incorporate prior. Kundu\cite{kundu2022panoptic} uses meta-learning to find a good category-specific initialization and employ instance-specific fully weight encoded functions to represent each object in scene. Most of the previous work is based on dense point clouds or images to complete instance reconstruction, Boulch\cite{boulch2021needrop} adopted sampling strategy of picking needles with end points on opposite sides or on the same side of the surface to realize dense reconstruction with sparse point cloud, this method requires pre-training on the dataset where sparse point clouds reside.

\begin{figure*}[!t]
\centering
\includegraphics[width=0.95\textwidth]{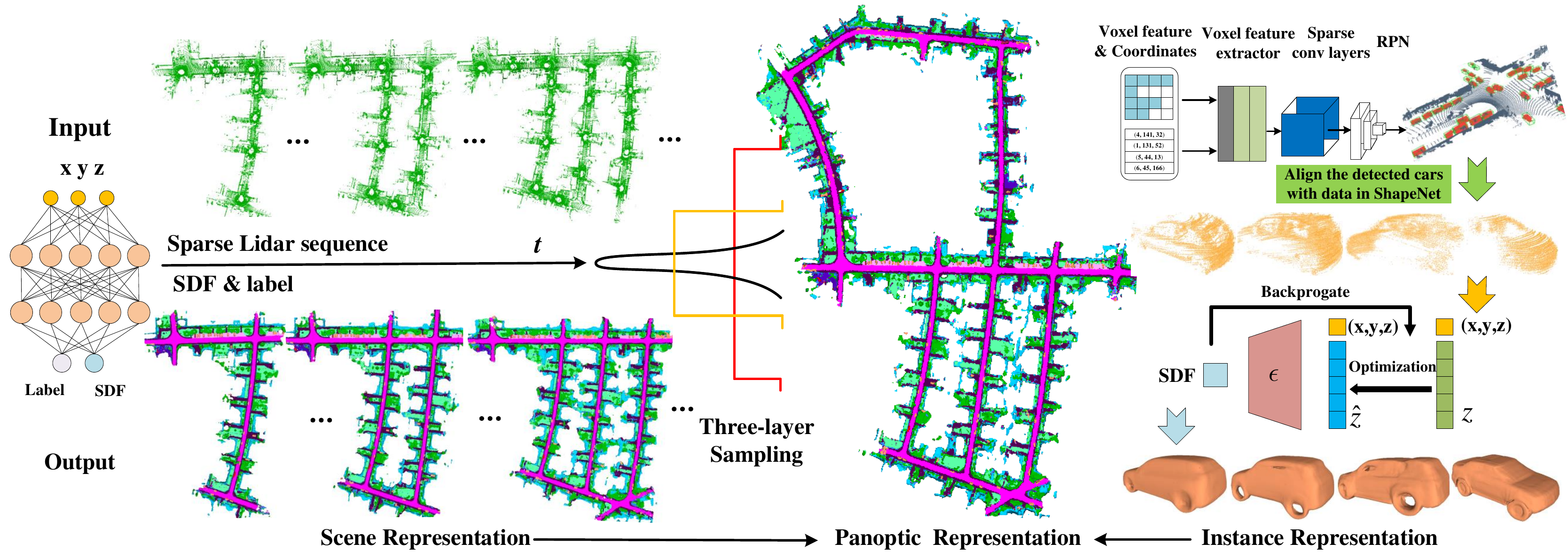}%
\caption{Given sequential sparse data, our model continuously learns scene property with three-layer sampling strategy that covers different level information including global, local and near-surface to achieve implicit semantic scene representation. In addition, we pre-train a category-specific MLP as prior to complete dense reconstruction of vehicles even with serious data default.}
\label{fig_sim}
\vspace{-4mm}
\end{figure*}

\section{Formulation}
\label{formulation}
The panoptic representation $\mathcal{S}$ comprises two subcomponents, namely ${^S\mathcal{S}}$ and ${^I\mathcal{S}}$, which respectively capture information pertaining to the scene and the instances within it. The former is updated continuously by means of the LiDAR stream, while the latter is constructed based on detected instances in conjunction with category-specific priors.

\textbf{Scene:} We are committed to modelling 3D environment continually through LiDAR streams $D^t$ with an implicit representation $\mathcal{F}(\cdot)$. The $D^t=(\boldsymbol{p}_{i}^{t}, \boldsymbol{s}_{i}^{t})$ is composed of point cloud coordinates $\boldsymbol{p}_{i}^{t}$ and corresponding attributes $\boldsymbol{s}_{i}^{t}$. 
\begin{equation}
   \boldsymbol{s}_{i}^{t}=\mathcal{F}(\boldsymbol{p}_{i}^{t};\theta^{t})
\end{equation}
Here, $t$ is time stamp, and $\boldsymbol{s}$ is property of scene that is represented by the SDF whose sign indicates whether the region is inside (-) or outside (+) of the shape. $\mathcal{F}(\cdot)$ is a continuous function that maps spatial point $\boldsymbol{p}$ to its distance to the nearest boundary, and the surface ${^S\mathcal{S}}$ of the scene is represented by the iso-surface of $\mathcal{F}(\cdot)=0$:
\begin{equation}
  ^S\mathcal{S}=\left\{\boldsymbol{p}_{i}^{t} \in \mathbb{R}^{3} \mid \mathcal{F}\left(\boldsymbol{p}_{i}^{t} ; \theta^{t}\right)=0\right\}, \mathcal{F}(\cdot): \mathbb{R}^{3} \mapsto \mathbb{R}.  
\end{equation}
According to the properties of SDF\cite{ortiz2022isdf}, it is to solve a specific Eikonal boundary value problem which restricts the norm of spatial gradients $\nabla_{\boldsymbol{p}} \mathcal{F}$ approach to 1 almost everywhere: $\left|\nabla_{\boldsymbol{p}} \mathcal{F}(\boldsymbol{p})\right|=1$. Morever, if points are sufficiently close to the surface, the $\mathcal{F}(\boldsymbol{p})$ is espected to be $0$, and the gradient $\nabla_{\boldsymbol{p}}\mathcal{F}$ is equal to the surface normal:  $\nabla_{\boldsymbol{p}} \mathcal{F}(\boldsymbol{p})=\mathbf{n}(\boldsymbol{p})$.
Therefore, we fit a neural network $\mathcal{F}(\cdot)$ that parameterize $\boldsymbol{s}$ to model a scene $^S\mathcal{S}$ from LiDAR streams $D^t$ using a loss of the form:
\begin{equation}
\label{loss}
\begin{aligned}
\mathcal{L}_{\mathrm{sdf}}&=\int_{\Omega}\left\|\left|\boldsymbol{\nabla}_{\boldsymbol{p}} \mathcal{F}(\boldsymbol{p})\right|-1\right\| d \boldsymbol{p}\\
&+\int_{\Omega_{0}}\|\mathcal{F}(\boldsymbol{p})\|+\left(1-\left\langle\boldsymbol{\nabla}_{\boldsymbol{p}} \mathcal{F}(\boldsymbol{p}), \mathbf{n}(\boldsymbol{p})\right\rangle\right) d\boldsymbol{p}\\
\end{aligned}
\end{equation}
The $\Omega$ denotes the entirety of the domain, wherein the zero-level set of the SDF is represented as $\Omega_0$. Additionally, to remedy the lack of constraints on off-surface points, another constraint is introduced:
\begin{equation}
   \mathcal{L}_\mathrm{off}= \int_{\Omega \backslash \Omega_{0}} \psi(\mathcal{F}(\boldsymbol{p})) d \boldsymbol{p}
\end{equation}
Here, $\psi(\boldsymbol{p})=\exp (-\alpha \cdot|\mathcal{F}(\boldsymbol{p})|)$, $\alpha \gg 1$ penalizes off-surface points for creating SDF values close to 0. Generally, we assigns a value of -1 to all points off the surface, resulting in a consistent contribution for fitting the surface. In the context of mapping large-scale scenes, the effectiveness of \textbf{global uniform sampling} may be compromised by the significant scale variations. While global uniform sampling helps to avoid the random prediction of SDF values in free space, the majority of the sampled points are typically invalid with negligible impact on the scene surface during the initial stages of map update. To expedite the fitting of function  $\mathcal{F}(\cdot)$, \textbf{local sampling} is performed by leveraging information from the current local map. However, with increasing iterations, the neural network may suffer from a loss of pertinent geometric information in the vicinity of the surface. In response, we propose the adoption of a \textbf{near-surface sampling} method to address this challenge. As such, the proposed methodology involves sampling off-surface points at three different levels, characterized by sampling proportions $\lambda_g$, $\lambda_l$ and $\lambda_n$ for global, local, and near-surface sampling, respectively, such that the sum of the proportions equals 1: $\lambda_g+\lambda_l+\lambda_n=1$.

\textbf{Instance:} 
We denote a shape of instance ${^I\mathcal{S}}:\{\boldsymbol{p_1},\boldsymbol{p_2},\cdots,\boldsymbol{p_j}\}$ with a signed distance function $f_{\epsilon}$:
\begin{equation}
    {^I\mathcal{S}}:=\{(\boldsymbol{p_j}, s_j): f_{\epsilon}(\boldsymbol{p_j})=s_j\}
\end{equation}
Here, $\boldsymbol{p_j} \in \mathbb{R}^{3}$ are coordinates of the shape. 
Inspired by DeepSDF\cite{park2019deepsdf}, an auto-decoder neural network $f_{\epsilon}(\cdot)$ is trained in a large amount of homogeneous instances with diverse shapes, which contains common properties of this class. Additionally, we introduce a latent vector $\boldsymbol{z}$, which can be thought of as encoding the desired shape. Given a sparse or partial shape, we will adopt a probabilistic perspective to derive the process of instance reconstruction. The posterior over shape code $\boldsymbol{z}$ which is paired with observed shape ${^I\mathcal{S}}$ can be decomposed as:
\begin{equation}
    p\left(\boldsymbol{z} \mid {^I\mathcal{S}}\right)=p\left({^I\mathcal{S}}\right) \prod_{\left(\boldsymbol{p}_{j}, \boldsymbol{s}_{j}\right) \in {^I\mathcal{S}}} p_{\epsilon}\left(\boldsymbol{s}_{j} \mid \boldsymbol{z} ; \boldsymbol{p}_{j}\right)
\end{equation}
where $\epsilon$ parameterizes SDF prior, and $ p_{\epsilon}\left(\boldsymbol{s}_{j} \mid \boldsymbol{z} ; \boldsymbol{p}_{j}\right)$ is expressed via a deep feed-forward network $f_{\epsilon}\left(\boldsymbol{z}, \boldsymbol{p}_{j}\right)$:
\begin{equation}
    p_\epsilon\left(\boldsymbol{s}_j \mid \boldsymbol{z} ; \boldsymbol{p}_j\right)=\exp \left(-\mathcal{L}_{\rm{ins}}\left(f_\epsilon\left(\boldsymbol{z}, \boldsymbol{p}_j\right), s_j\right)\right)
    \label{prior}
\end{equation}
The loss function $\mathcal{L}_{\rm{ins}}$ serves to penalize deviations between predicted and actual signed distance function (SDF) values, $s_j$. Instances denoted by ${^I{\mathcal{S}}}$, where SDF values of points are constrained to zero, is interpreted as likelihoods. During optimization, we maximize the joint log posterior over the reconstructing shape to obtain the shape code $\boldsymbol{z}$:
\begin{equation}
\hat{\boldsymbol{z}}=\underset{\boldsymbol{z}}{\arg \min } \sum_{\left(\boldsymbol{p}_{j}, \boldsymbol{s}_{j}\right) \in {^I\mathcal{S}}} \mathcal{L}_{\rm{ins}}\left(f_{\epsilon}\left(\boldsymbol{z}, \boldsymbol{p}_{j}\right), s_{j}\right)+\frac{1}{\sigma^{2}}\|\boldsymbol{z}\|_{2}^{2}
\end{equation}
We assume the latent shape-code space submits to a zero-mean multivariate-Gaussian distribution with
a spherical covariance $\sigma^{2}I$. Finally, the $\hat{\boldsymbol{z}}$ are concatenated with relative coordinates, and then feed it into the trained neural network to inference SDF values of reconstructing shape.

\section{Method}


\subsection{Network architecture}
Following the network architecture in SIREN\cite{sitzmann2020implicit}, We model the SDF $s$ using an MLP with 4 hidden layers of feature size 256, map a 3D coordinate $\boldsymbol{p}=(x,y,z)$ to a SDF value: $\mathcal{F}(\boldsymbol{p};{\theta}) = s$.  Fourier Feature Networks\cite{tancik2020fourier} transform the effective neural tangent kernel (NTK) into a stationary kernel with a tunable bandwidth applying Bochner’s theorem. We use a 
fourier feature mapping to $\gamma(\boldsymbol{p})=[\cos (2 \pi \mathbf{B}\boldsymbol{p}), \sin (2 \pi \mathbf{B} \boldsymbol{p})]^{\mathrm{T}}$, where each entry in $\mathbf{B} \in \mathbb{R}^{m \times d}$ is sampled from $\mathcal{N}\left(0, \sigma^{2}\right)$, and $\sigma$ is chosen for each task and dataset with a hyperparameter sweep. In the absence of any strong prior on the frequency spectrum of the signal, we use an isotropic Gaussian distribution:
\begin{equation}
\label{fourier mapping}
\begin{aligned}
\gamma(\boldsymbol{p})=[a_{1} \cos \left(2 \pi \mathbf{b}_{1}^{\mathrm{T}} \boldsymbol{p}\right), a_{1} \sin \left(2 \pi \mathbf{b}_{1}^{\mathrm{T}} \boldsymbol{p}\right), \ldots,\\ a_{m} \cos \left(2 \pi \mathbf{b}_{m}^{\mathrm{T}} \boldsymbol{p}\right),a_{m} \sin \left(2 \pi \mathbf{b}_{m}^{\mathrm{T}} \boldsymbol{p}\right)]^{\mathrm{T}}
\end{aligned}
\end{equation}
In addition, we add a network with the same structure in parallel to output the semantic value of each point. The architectural framework utilized for the purpose of instance reconstruction is based on the DeepSDF\cite{park2019deepsdf} paradigm.

\subsection{Sampling}
\label{sampling}
As delineated in Section \ref{formulation}, the $\mathcal{L}_\mathrm{off}$ can be cast in loss functions to penalize deviations, where the off-surface point constraints are represented by $\psi(\mathcal{F}(\boldsymbol{p}))$. 
To accommodate new data streams, we adopt a practical approach by sampling $D^t$ over the entire domain $\Omega$, which includes both on-surface points $\Omega_0$ whose SDF values are 0, as well as off-surface points $\Omega \backslash \Omega_{0}$, with SDF values set to -1.

\textbf{On-surface sampling:} To make a trade-off between point cloud density and computing efficiency, we adopt a strategy of selecting key frames at regular intervals of three frames and utilize them to update the neural network via learning. To prevent catastrophic forgetting of past scenes, we randomly sample 75\% of points from the previous key frames and the remaining 25\% from the latest key frames. However, due to the presence of overly redundant samples of buildings and roads, as well as a decrease in the proportion of small instances as the scene size increases, the surface fitting of small instances may be inadequate. To address this issue, we employ importance sampling based on semantic information to ensure a relevant instance sampling scale. Specifically, at each iteration, we extract $N_g$ on-surface points within a vicinity of $n_o$ points of instances. In this study, $N_g$ and $n_o$ are set to 140000 and 6000, respectively.

\begin{figure}[!t]
\centering
\includegraphics[width=0.45\textwidth]{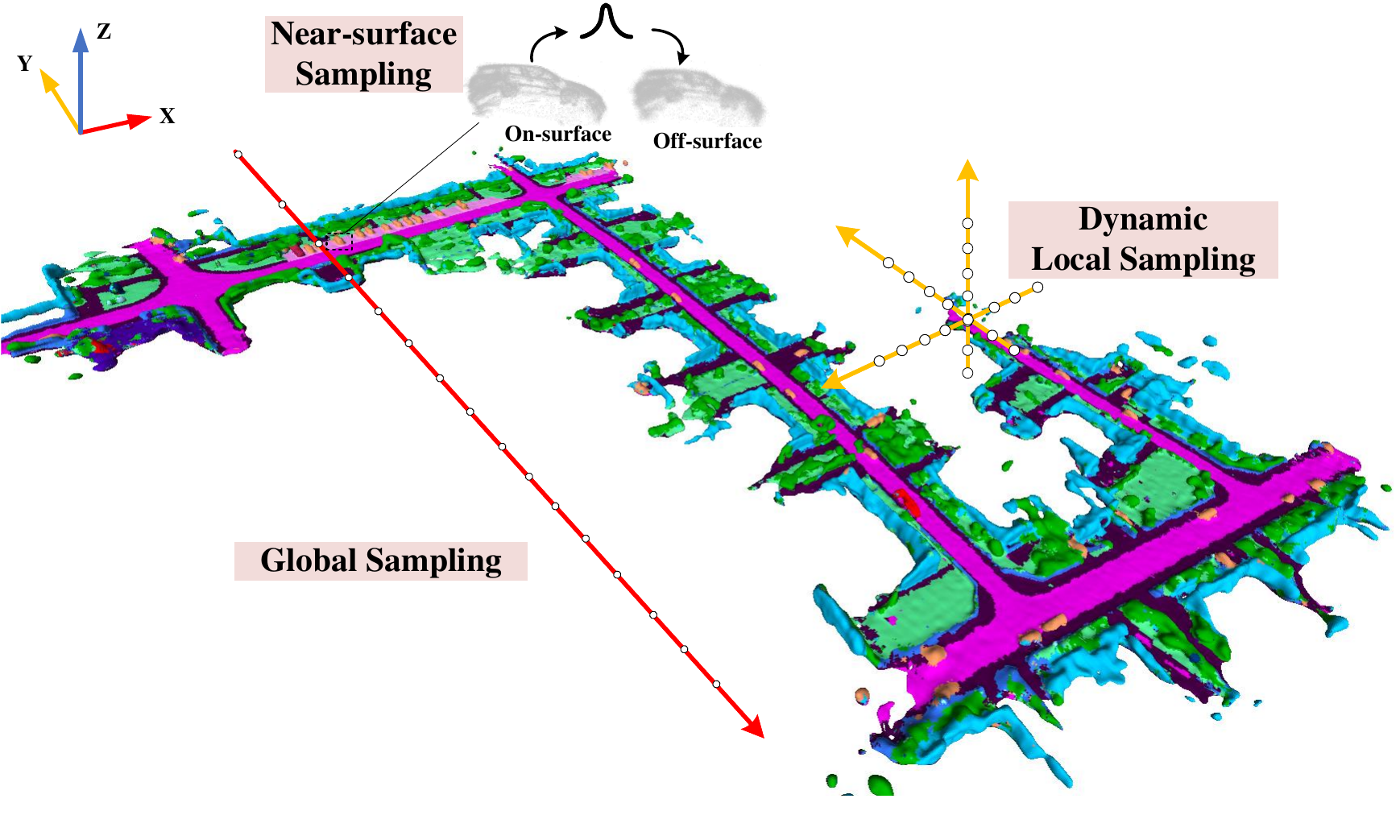}%
\caption{Three-layer sampling that covers global, dynamic local and near-surface. 
The global sampling is to avoid the random prediction in free space, and the local sampling is to capture local change to facilitate fitting of neural network during continue learning, and near-surface sampling is attached to mitigate the forgetting of detail. 
}
\label{threesampling}
\vspace{-4mm}
\end{figure}

\textbf{Off-surface Three-layer Sampling:}
The present study concerns a key challenge in global uniformly sampling, where the majority of points fall in free space, causing a loss of surface information due to uniform penalty assignment. Specifically, as the SDF values are uniformly set to -1 for all points off the surface, points near the surface receive the same penalty as points far away from the surface.
To overcome this limitation, a novel three-layer sampling strategy has been proposed, as illustrated in Fig.\ref{threesampling}. This strategy leverages the complementary strengths of different sampling methods by incorporating three levels of information, namely global, local, and near-surface sampling, each with a different proportion $\lambda_g$, $\lambda_l$ and $\lambda_n$.

For a limited scene or an individual object, the point cloud's scale in all directions remains close, and off-surface points obtained through global uniform sampling are adequate to represent different levels of noise. However, in an urban scene where the map continually updates, the scale difference in all directions gradually intensifies. Consequently, the scale map in the $\mathbf{Z}$ direction becomes almost negligible compared to the scale of the scene in the $\mathbf{X}$ and $\mathbf{Y}$ directions. Hence, uniform sampling of the entire space alone is insufficient to represent the noise of local geometry. The results depicted in Fig. \ref{ablation} demonstrate that the network trained exclusively on global sampling mainly learned the overall outline information of the scene. Nevertheless, the overall contour information remains vital, and thus a certain proportion of points, denoted by $\boldsymbol{p_g}$, acquired through global sampling is essential:
\begin{equation}
\begin{aligned}
    &\boldsymbol{p_g} = \mathcal{U}([-1,-1,-1],[1,1,1]),\\
    &\lambda_g=size(\boldsymbol{p_g})/size(N_g).
\end{aligned}
\end{equation}


If the continual update employs global uniform sampling, it may result in negligible contributions from most off-surface points at the onset, leading to the loss of local geometric details. In response to the influx of new LiDAR stream, the dynamic boundary ($\boldsymbol{b_l}$ and $\boldsymbol{b_u}$) of the scene is estimated, where $\boldsymbol{b_l}$ denotes the lower limit and $\boldsymbol{b_u}$ denotes the upper limit of the scene boundary. Subsequently, local off-surface points $\boldsymbol{p_l}$ are uniformly sampled within this range to maximize the network's ability to capture the local scene changes, $\boldsymbol{p_l} \sim \mathcal{U}(\boldsymbol{b_l},\boldsymbol{b_u})$.
\begin{equation}
\begin{aligned}
&\boldsymbol{b_l}=(\frac{\boldsymbol{L_{min}} -{G_{min}}}{{G_{max}}-{G_{min}}} - 0.5)\times 2\\
&\boldsymbol{b_u}=(\frac{\boldsymbol{L_{max}} -{G_{min}}}{{G_{max}}-{G_{min}}} - 0.5)\times 2\\
\end{aligned}
\end{equation}
 Here, the maximum and minimum coordinate vectors of the extant local point clouds are denoted by $\boldsymbol{L_{max}} (x_{max}^l,y_{max}^l,z_{max}^l)$ and $\boldsymbol{L_{min}} (x_{min}^l,y_{min}^l,z_{min}^l)$, respectively, both of which adjust dynamically to accommodate the influx of new data. ${G_{max}}$ and ${G_{min}}$ are the maximum and minimum coordinate values of the entire scene. However, local dynamic sampling in isolation is inadequate, owing to the fact that off-surface points lying outside the local scene do not contribute to the network, which makes the neural network predict random SDF values in free space.
Therefore, a certain proportion of local and global sampling are both necessary.

\begin{figure}[h]
\centering
\includegraphics[width=0.45\textwidth]{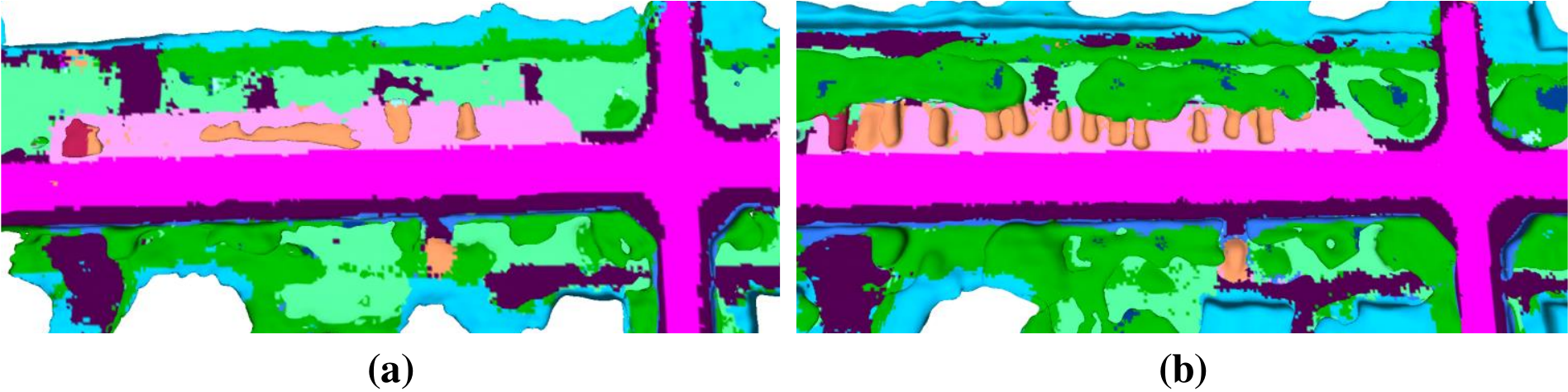}%
\caption{(a) Without near-surface sampling, the network gradually forgets the information of instances like cars and trees in the scene. (b) Near-surface sampling successfully mitigate catastrophic forgetting of instances.}
\label{forget}
\vspace{-5mm}
\end{figure}

The aforementioned sampling strategies provide a comprehensive and adaptable depiction of the entire and local scene,  However, as the volume of data increases, the effects of local sampling become increasingly similar to those of global sampling, with the exception of the Z dimension's spatial scale. Consequently, the neural network gradually loses sight of the surface information of the scene, which is illustrated in Figure \ref{forget}.

As a result, off-surface points $\boldsymbol{p_n}$ close to the surface of scene $^S\mathcal{S}$ are sampled to aggravate the penalty of noise near the surface in the learning process:
\begin{equation}
    \boldsymbol{p_n}=\left\{(\boldsymbol{p}+\mathbf{h}, \boldsymbol{p}-\mathbf{h}) \mid \boldsymbol{p} \in \mathcal{S}, \mathbf{h} \sim \mathcal{N}\left(0, \sigma_{h}\right)\right\}
\end{equation}
where $\mathbf{h}$ is randomly sampled from the multivariate Gaussian distribution $\mathcal{N}\left(0, \sigma_{h}\right) \in \mathbb{R}^{3}$ with standard deviation $\sigma_{h}$, $\sigma_{h}$ is $diag(0.0003)$ in this paper.

\subsection{Instance Representation}
In the scene reconstruction, the outcomes of the instances exhibit a deficiency of intricate particulars. As an illustration, vehicles appear as convex components on the map, as depicted in Fig.\ref{forget}(b). One of the core benefits of an object-aware approach is able to incorporate inductive bias that objects instances within the same category often have similar 3D shapes and appearance. Taking this as inspiration, we introduce a category-specific prior by sharing a neural network across the object instances to well depict details of instances with absent data in the map. 
Taking the car as an example,  
\begin{figure}[h]
    \centering
    \includegraphics[width=0.41\textwidth]{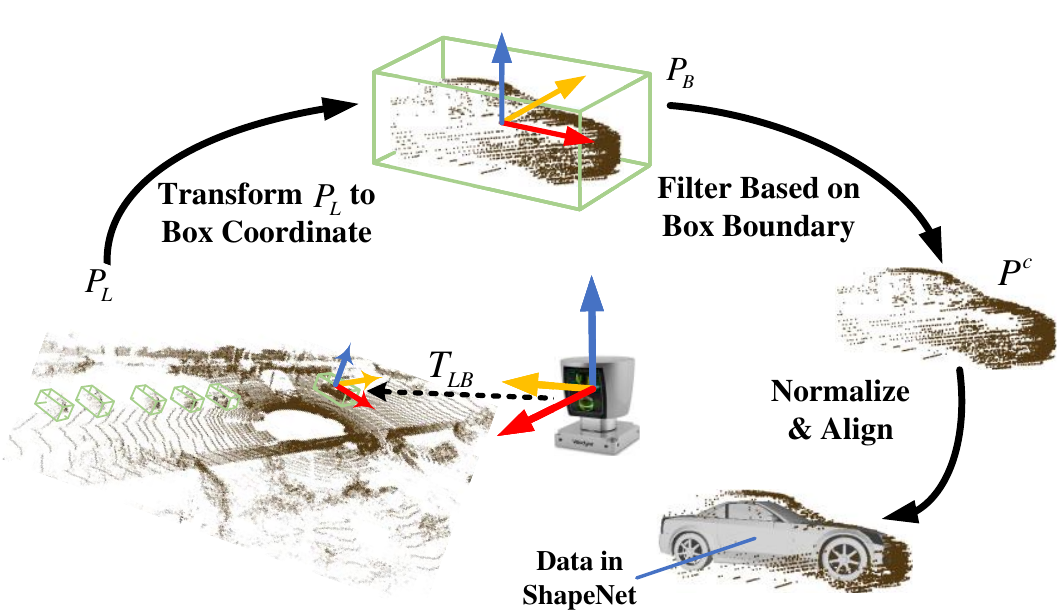}
    \caption{Align the detected cars with data in ShapeNet\cite{chang2015shapenet}.}
    \label{fig:align}
    \vspace{-2mm}
\end{figure}
we first use SECOND\cite{yan2018second} to detect the bounding box of cars, and transfer the corresponding point cloud from the LiDAR coordinate system to the central coordinate system of the bounding box, $P_B={T_{LB}}^{-1} \cdot P_L$, where $T_{LB}$ is transformation of the bounding box center with respect to the LiDAR coordinate system, $P_L$ is the raw LiDAR data, $P_B$ is the coordinate relative to bounding box center coordinate system. Then we extract the vehicle point cloud $P^c$ from $P_B$ according to the boundary of bounding box. Next, normalize $P^c$ to [-0.5,0.5] with the range of bounding box as the maximum and minimum value and align with the data in ShapeNet\cite{chang2015shapenet}, as shown in Fig.\ref{fig:align}.
What's next, following the depiction in section \ref{formulation} (Fig.\ref{fig_sim}),
given fixing $\epsilon$ trained in cars of ShapeNet by DeepSDF\cite{park2019deepsdf}, a latent code  $\hat{\boldsymbol{z}_i}$ for vehicle $P^{c}_{i}$ can be estimated via MAP estimation. We 
concatenate the $\hat{\boldsymbol{z}_i}$ with relative coordinates of vehicle, and then feed them into the MLP of DeepSDF. Adam optimizer is used to update the latent code, and complete the reconstruction of the vehicle instance via Marching Cube according to the SDF value predicted by the network. Finally, the vehicle is scaled and converted to the scene map coordinate system according to the pose of the center relative to bounding box, the panoptic scene is represented in Fig.\ref{fig:teaser}(e). 
\subsection{Training and Inference}
\label{training}
\textbf{Training:} 
For scene representation, besides the geometry constraints described in section \ref{formulation}, we also add a parallel implicit generative
head to directly model the implicit semantic label field. Its
structure is similar to our SDF model, except that
it outputs the probabilities of label classification.
We supervise the semantic segmentation results with a multi-classification cross entropy loss:
\begin{equation}
\mathcal{L}_{\rm{seg}} = -{\frac 1 {N_{g}}}\sum_{i=1} ^ {N_{g}}\sum_{c=1} ^ {C} y_{i,c} {\rm{log}}(pr_{i,c}).
\end{equation}
where $y_{i,c}$ and $pr_{i,c}$ are the actual and predicted probability for point $i$ belonging to category $c$ respectively. As per the prescribed constraints, $N_g$ points are sampled at random from $\Omega_{0}$, while an equal number of off-surface points are selected from $\Omega \backslash \Omega_{0}$, employing our three-layer sampling strategy that involves the selection of $\lambda_g$, $\lambda_l$ and $\lambda_n$ in the proportions of 0.55, 0.35 and 0.1, respectively. The optimization of the scene representation is executed by minimizing the loss function:
\begin{equation}
\mathcal{L}=\mathcal{L}_{\mathrm{sdf}}+\mathcal{L}_{\mathrm{off}}+\mathcal{L}_{\rm{seg}} 
\label{loss}
\end{equation}

Regarding instance representation, given a set of cars of ShapeNet, we train a category-shared MLP $f_{\epsilon}$, following the DeepSDF approach, whereby we minimize the aggregate of losses between the predicted and actual SDF values of the points within the cars, subject to the following loss function:
\begin{equation}
    \mathcal{L}_{\rm{ins}}\left(f_\epsilon(\boldsymbol{p}), s\right)=\left|\operatorname{clamp}\left(f_\epsilon(\boldsymbol{p}), \delta\right)-\operatorname{clamp}(s, \delta)\right|,
\end{equation}
where $\operatorname{clamp}(\boldsymbol{p}, \delta):=\min (\delta, \max (-\delta, \boldsymbol{p}))$ introduces the
parameter $\delta$ to control the distance from the surface over which we expect to maintain a metric SDF. 

\textbf{Inference:} Given the SDF values and semantic labels by our scene representation, semantic mapping is performed. For city scene, scale in Z direction is almost negligible comparing with the scales in X and Y. When inference, sampling same number of points in all directions like the Marching Cube\cite{lorensen1987marching} will lead to large amount of invalid samples in Z-axis, which gives rise to insufficiency of memory usage. In view of this, we will sample $N_x \times N_y \times N_z$ points, where $N_{x}=N_{y}$, and the $N_z$ is:
\begin{equation}
N_z=\frac{{Z_{max}}-{Z_{min}}}{{G_{max}}-{G_{min}}}\times N_{x}.
\end{equation}
Here, ${Z_{max}}$ and ${Z_{min}}$ are the maximum and minimum values of the map on the Z-axis respectively, $G_{max}$ and $G_{min}$ are the maximum and minimum values of the map in all directions. We sample points on the Z-axis from the position $Z_{start}$ where there is information instead of 0, and the $Z_{start}$ is:
\begin{equation}
\label{zstart}
Z_{start}=\frac{Z_{min}-G_{min}}{G_{max}-G_{min}}\times N_{x}.
\end{equation}
As a consequence, we can  guarantee optimal  utilization of memory resources while generating high-resolution meshes. 
Simultaneously, we will render the produced mesh in color, with the color of each vertex determined by the label of its closest spatial point.

For the representation of instance, given a normalised car detected from LiDAR sequence, we firstly feed the concatenated vector including coordinates and initialized latent code to the $f_{\epsilon}$, and optimize the latent code through back-propagation. The coordinates and corresponding latent code with fixed parameters are cascaded into the trained neural network, the SDF values are output and the Marching Cube\cite{lorensen1987marching} is used to generate corresponding mesh.

\section{Experiments and Results}
\begin{figure}[!t]
\centering
\includegraphics[width=0.45\textwidth]{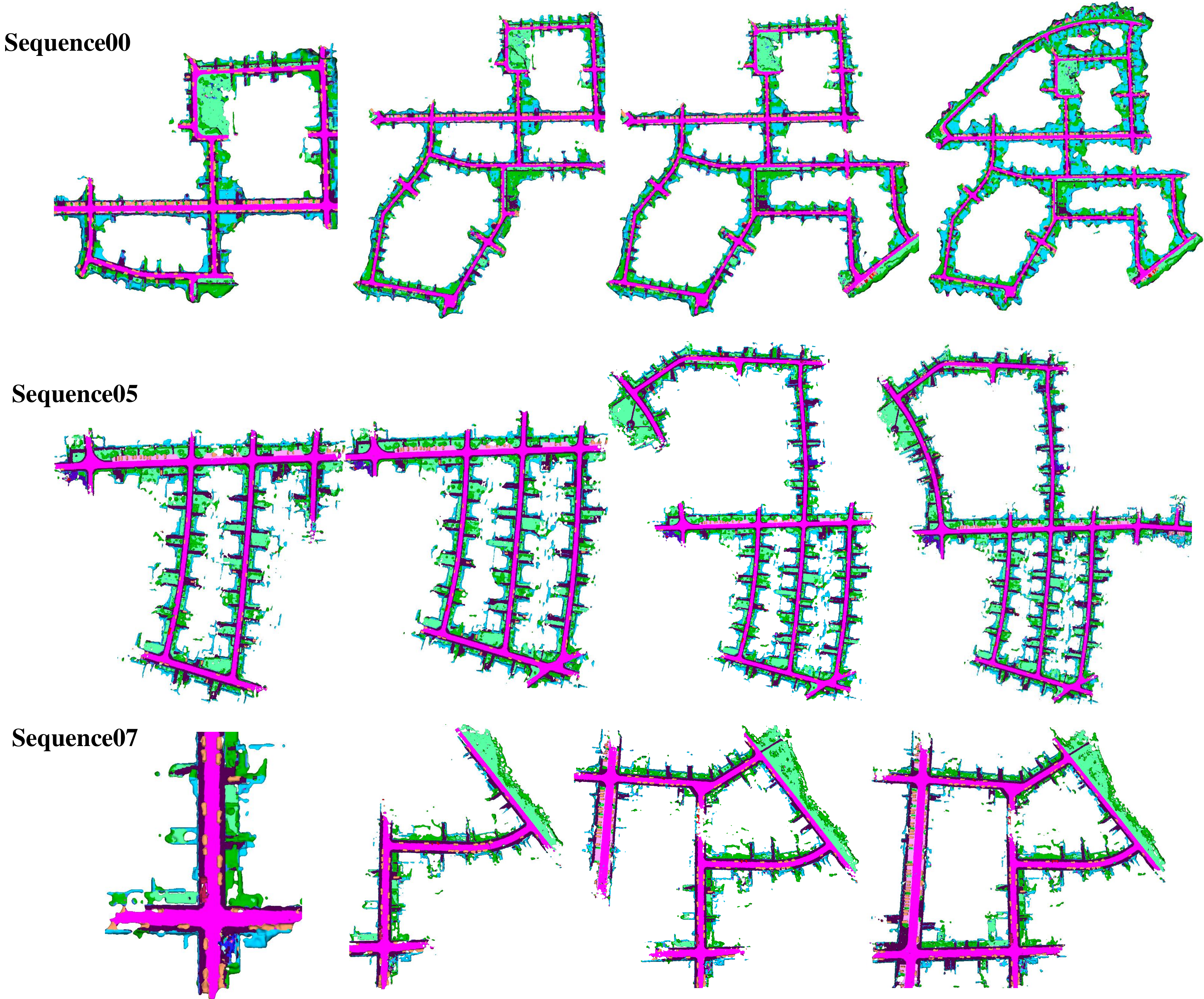}%
\caption{We select three LiDAR odometry sequences of different sizes from the SemanticKITTI: small (07), medium (05), and large (00) to execute continual implicit semantic mapping. }
\label{07}
\vspace{-4mm}
\end{figure}

We evaluate our method on three city-scale sequences of  SemanticKITTI\cite{semantickitti} odometry at large, medium and small magnitudes: 00 (4541 scans), 05 (2761 scans) and 07 (1101 scans), and visualisation results are shown in Fig.\ref{07}. All experiments are conducted on a Linux system with Intel Core i9-12900K CPU at 5.2GHz, and an NVIDIA GeForce RTX A4000 GPU with 16GB of memory.

\subsection{Data and Metrics}
\textbf{Data Preparation:}  Before training, the outliers and dynamic information are deleted in line with the ground truth of semantic label.
In addition, to obtain the prior for our cars, we trained the category-specific shared MLP on cars of ShapeNet\cite{chang2015shapenet}. 
The reconstructing cars of every frame are extracted and normalized to [-0.5, 0.5], as described in section IV(C). Owing to a slight deviation in the pose of data obtained by the 3D detection network, we fine tune the angle to align our cars as closely as possible with the car in ShapeNet\cite{chang2015shapenet}.

\textbf{Metrics:} 
We evaluate both the reconstruction quality and semantic segmentation of the system. For reconstruction quality evaluation, we uniformly sample 1,000,000 points from the ground-truth points and reconstructed meshes, respectively, then report the following metrics. \textit{Chamfer Distance} (denoted as CD in Tab.\ref{tab:sampling} \ref{tab:PosEncoding} \ref{tab:qu_rec}) finds the nearest point in the other point set, and sums the square of distance up. \textit{Precision} is the fraction of the points from reconstructed mesh that is closer to points in the ground truth than a threshold distance, which is set to 0.5m. \textit{Recall} refers to the fraction of the ground truth points closer to the points in the reconstruction mesh than 0.5m. \textit{Fscore} is the average of accuracy and completeness and is used to quantify the overall reconstruction quality.  Besides, the \textit{mean interactions over union (mIoU)} are used in semantic segmentation evaluation. 

\begin{figure}[h]
\centering
\includegraphics[width=0.5\textwidth]{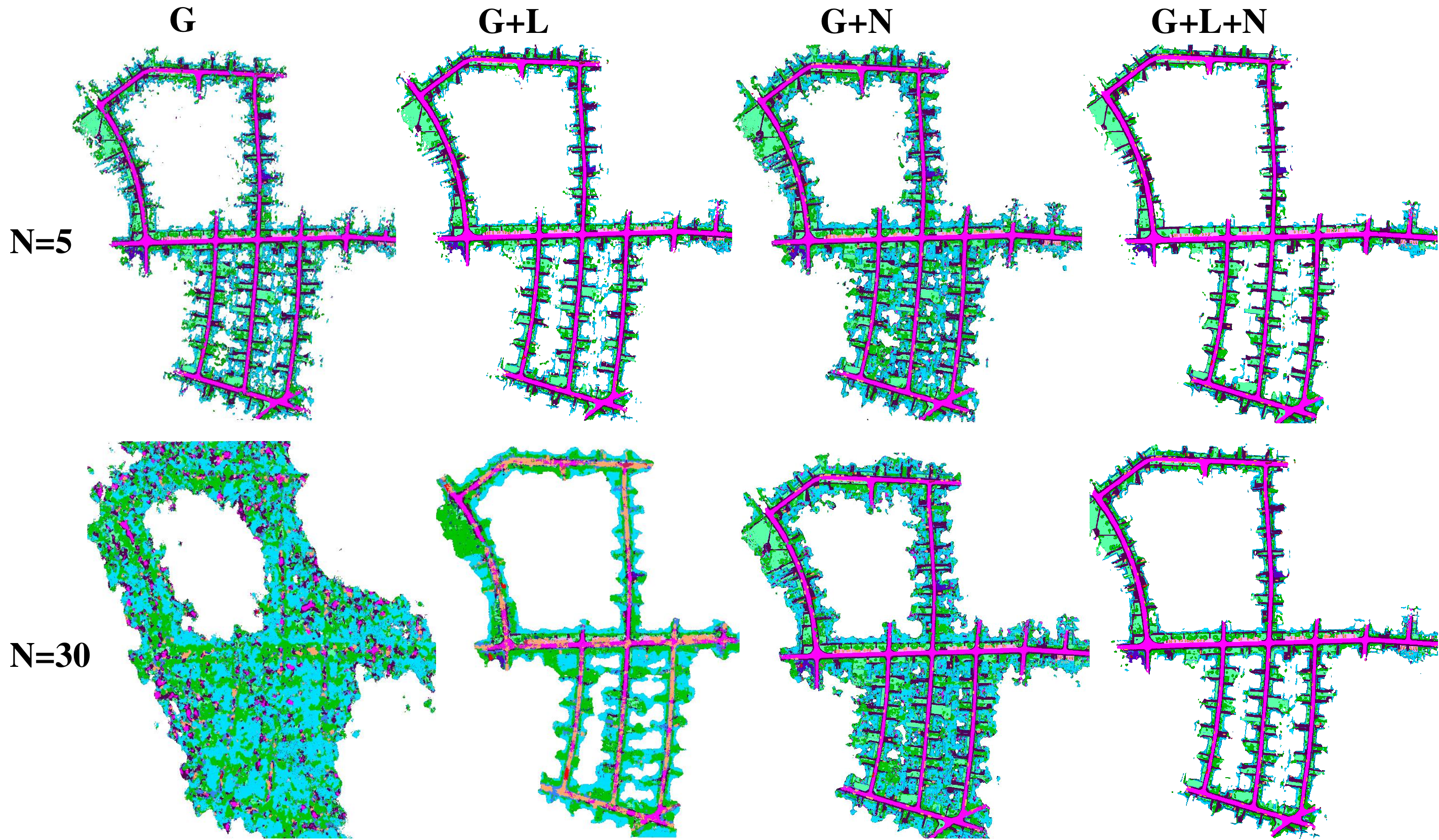}%
\caption{Ablation study for sampling strategy with different number of iteration. We denote N as iteration number, G, L and N represent the global, local and near-surface sampling, respectively. 
}
\label{ablation}
\vspace{0mm}
\end{figure}

\begin{table}[htb]
\centering
\caption{Ablation study for Three-layer Sampling strategy.}
\renewcommand\arraystretch{1.0}
\resizebox{0.98\linewidth}{!}{
\begin{tabular}{cccccc}
\toprule
\begin{tabular}[c]{@{}c@{}} Iteration\\(epochs) \end{tabular}&
\begin{tabular}[c]{@{}c@{}}\makecell[c]{Global\\ Sampling}\end{tabular} & \begin{tabular}[c]{@{}c@{}}Local\\ Sampling \end{tabular}& 
\begin{tabular}[c]{@{}c@{}} Near-surface\\ Sampling \end{tabular}& 
\begin{tabular}[c]{@{}c@{}} CD(m) \end{tabular}&
\begin{tabular}[c]{@{}c@{}} mIoU (\%)
\end{tabular}\\
\midrule
\multirow{4}*{5}&\checkmark&$\bm{\times}$&$\bm{\times}$&0.25&67.6\\
~&\checkmark&\checkmark&$\bm{\times}$&2.12&95.0\\
~&\checkmark&$\bm{\times}$&$\checkmark$&0.65&95.2\\
~&\checkmark&\checkmark&\checkmark&\textbf{0.09}&\textbf{96.6}\\
\midrule
\multirow{4}*{30}&\checkmark&$\bm{\times}$&$\bm{\times}$&{3.33}&72.4\\
~&\checkmark&\checkmark&$\bm{\times}$&4.55&74.9\\
~&\checkmark&$\bm{\times}$&$\checkmark$&1.00&96.1\\
~&\checkmark&\checkmark&\checkmark&\textbf{0.05}&\textbf{96.3}\\
\bottomrule
\end{tabular}
}
\label{tab:sampling}
\vspace{-4mm}
\end{table}

\subsection{Ablation study for three-layer sampling method}
\textbf{Sampling strategy and iteration number} play important roles in implicit reconstruction.
The reconstruction results in Fig.\ref{ablation} show that: With the increasing of iteration, adopting only global uniform sampling can easily lead to the forgetting of local geometry, meanwhile, the neural network learns some floating noise into the scene as well; On the basis, within global and local sampling, both reconstruction and semantic segmentation are improved in visualization, which alleviate the isolated point noise; After utilizing global and 
near-surface sampling, significant improvements in both reconstruction quality and semantic segmentation were observed in visualization and quantitative results. However, there was an issue of excessive local patch completion. Having employed the three-layer sampling method, the reconstruction and semantic segmentation achieve the best result. Quantitative evaluation results in Tab.\ref{tab:sampling} prove the effectiveness and necessity of the three-layer sampling method, which is robust to changes of iterations during training.

\begin{figure}[t]
\centering
\includegraphics[width=0.4\textwidth]{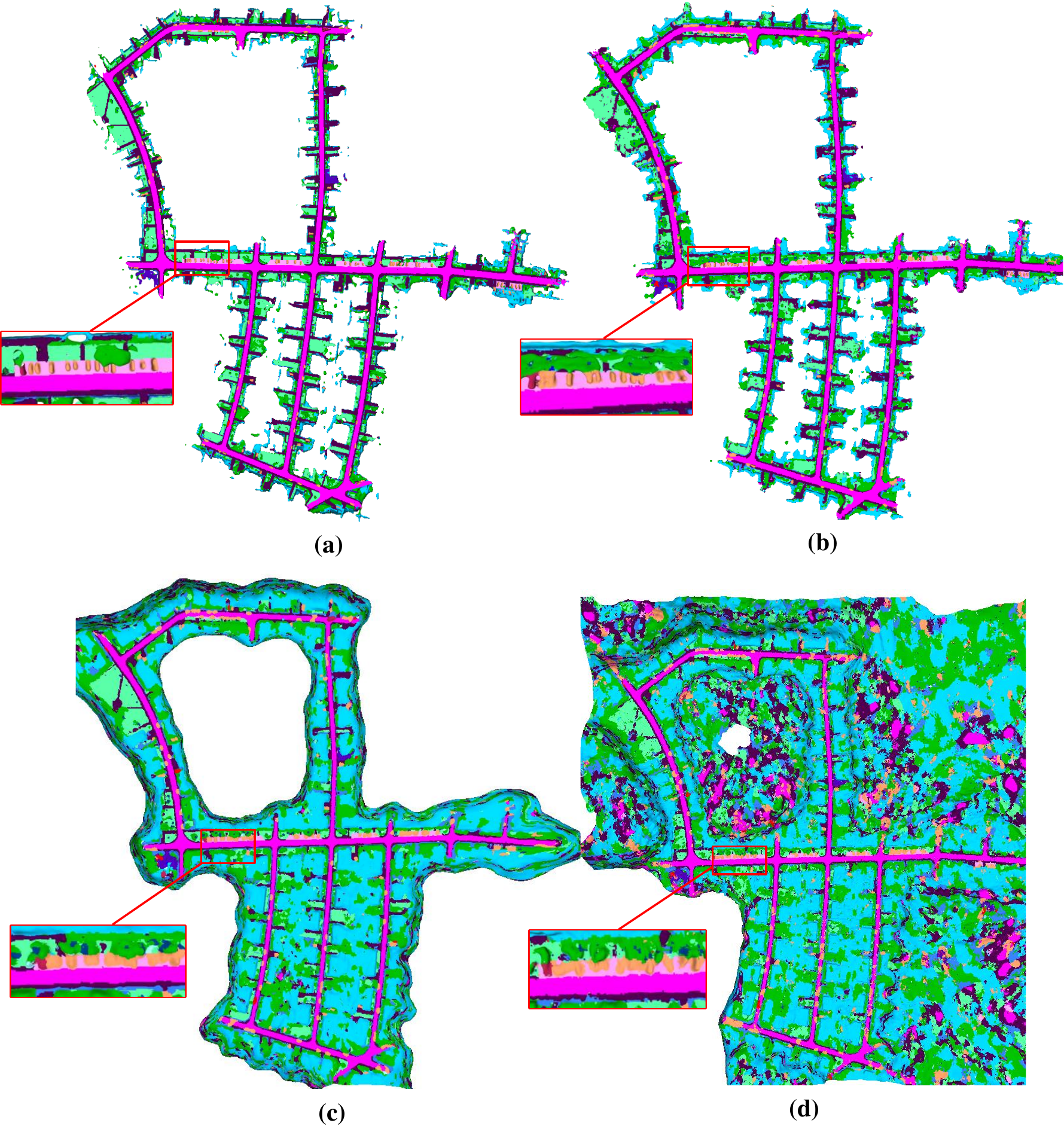}%
\caption{Representations with (a) Fourier encoding. (b) Positional encoding. (c) Learnable Fourier encoding. (d) no encoding}
\label{encodings}
\end{figure}

 \textbf{Encoding methods:} For coordinate-based MLPs, passing input points through a encoding method on a regression task is a prevailing practice. We compare the performance of our task with no input encoding and three encoding methods. One is Positional encoding that is consistent with the work proposed by Rahaman\cite{rahaman2019spectral} and its encoding level is 10, the other is Fourier encoding\cite{tancik2020fourier} with an isotropic Gaussian distribution used in this paper, and the last is Learnable Fourier (L-Fourier) encoding\cite{li2021learnable} that is the state-of-art method, as shown in Fig.\ref{encodings}.  Tab.\ref{tab:PosEncoding} shows that, in the case of using the three-layer sampling method, Fourier encoding performs best in semantic segmentation and reconstruction. 

\begin{table}[t]
\centering
\caption{Evaluation on reconstruction quality and segmentation of different encoding methods when the iteration number is 5.}
\renewcommand\arraystretch{1.0}
\begin{tabular}{lcc}
\toprule
\begin{tabular}[c]{@{}c@{}}\makecell[c]{Encoding\\ Methods}\end{tabular} & \begin{tabular}[c]{@{}c@{}} CD(m) \end{tabular}& mIoU(\%)\\
\midrule
Fourier &\textbf{0.09}&\textbf{96.6}\\
Positional &1.05&77.8\\
{L-Fourier}&7.17&90.0\\
No encoding &7.12&76.0\\
\bottomrule
\end{tabular}
\label{tab:PosEncoding}
\vspace{-2mm}
\end{table}


\subsection{Reconstruction Quality of Instance Representation} 
\begin{figure}[!t]
\centering
\includegraphics[width=0.45\textwidth]{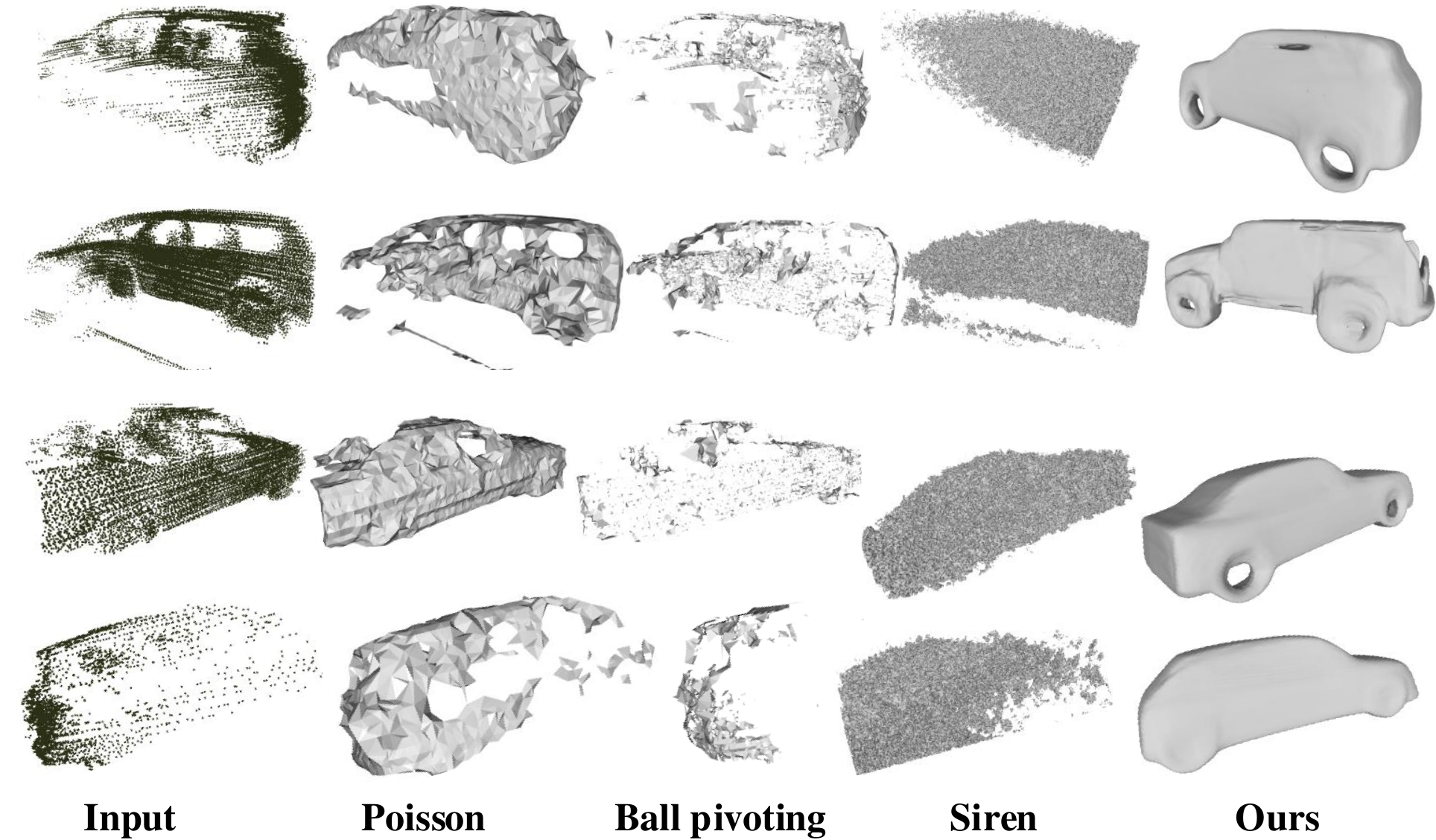}%
\caption{Qualitative comparison of our panoptic representation to other instance reconstruction  methods in KITTI.}
\label{car_com}
\end{figure}

We present a comparison of our method against two direct (non-learned) methods, Poisson meshing\cite{kazhdan2006poisson} and Ball Pivoting\cite{bernardini1999ball}, and a learning-based method SIREN\cite{sitzmann2020implicit}. The instances we will evaluate are obtained by the 3D detection algorithm without the complementary ground truth, so the reconstruction quality is only analyzed in terms of visualization (Fig.\ref{car_com}) and is not quantitatively evaluated. As expected, the direct methods fail to fit a proper shape let alone predict the missing part. While Poisson meshing  barely contains any detail, Ball Pivoting however
produces a detailed mesh around the input point cloud but it fails to reconstruct the hidden parts. For SIREN, once there are pieces of missing data, it is easy to cause under fitting and fail to reconstruct a complete instance. Inspired by the prior of category-specific prior, our method produce a realistic car
shape and reconstruct hidden parts of the cars, which outperforms the other methods.
\subsection{Comparison With Other Reconstruction Methods} 
We compared several reconstruction algorithm, including two explicit methods
Voxblox\cite{oleynikova2017voxblox} and VDBfusion\cite{vizzo2022sensors} 
and a implicit representation SIREN\cite{sitzmann2020implicit}. As shown in Tab.\ref{tab:map}, our model consists of scene model (1.8MB) and category-specific prior (2.4MB), which achieves the best performance in reconstruction (Tab.\ref{tab:qu_rec}), and the memory will not increase with the size of scene, as well as the SIREN with the minimum model memory (1.8MB). 
Conversely, the explicit representation not only exhibits a considerably high memory but also experiences a substantial growth with respect to the scale of the scene, rendering it impractical for very large scenes.
Our approach surpasses alternative explicit and implicit representation methodologies, and effectively make a good trade-off between reconstruction quality and memory efficiency.

\begin{table}[!t]
\centering
\caption{memory of different methods (MB)}
\renewcommand\arraystretch{1.0}
{
\begin{tabular}{ccccccc}
\toprule
\begin{tabular}[c]{@{}c@{}}\makecell[c]{Sequence}\end{tabular} &
\begin{tabular}[c]{@{}c@{}} Voxblox \end{tabular}&
\begin{tabular}[c]{@{}c@{}} VDBfusion \end{tabular}&
\begin{tabular}[c]{@{}c@{}} SIREN \end{tabular}&
\begin{tabular}[c]{@{}c@{}} Ours \end{tabular}\\
\midrule
00&448.9&3012.2&\textbf{1.8}&\textcolor[RGB]{155,0,0}{4.2}\\
05&264.6&2103.3&\textbf{1.8}&\textcolor[RGB]{155,0,0}{4.2}\\
07&82.4&748.1&\textbf{1.8}&\textcolor[RGB]{155,0,0}{4.2}\\
\bottomrule
\end{tabular}
}
\label{tab:map}
\end{table}

\begin{table}[!t]
\centering
\caption{Reconstruction Quality Evaluation on KITTI when the threshold is 0.5m }
\renewcommand\arraystretch{1.0}
\begin{tabular}{lcccccc}
\toprule
\begin{tabular}[c]{@{}c@{}}\makecell[c]{Methods}\end{tabular} &
\begin{tabular}[c]{@{}c@{}} CD(m) \end{tabular}&
\begin{tabular}[c]{@{}c@{}} Precision(\%) \end{tabular}&
\begin{tabular}[c]{@{}c@{}} Recall(\%) \end{tabular}&
\begin{tabular}[c]{@{}c@{}} Fscore(\%) \end{tabular}\\
\midrule
Voxblox&13.67&13.16&13.17&13.16\\
VDBFusion&1.72&22.31&30.69&25.84\\
SIREN&1.05&21.57&6.55&10.05\\
Ours&\textbf{0.05}&\textbf{96.81}&\textbf{70.90}&\textbf{81.86}\\
\bottomrule
\end{tabular}
\label{tab:qu_rec}
\vspace{-2mm}
\end{table}

\section{Conclusion}
We have presented a city-scale continual learning system with panoptic representation. For scene, when new LiDAR sreams come, three-layer sampling is adopted to ensure the global, local and approximate to surface information learning. For panoptic representation, a category-shared prior is pre-trained for implicit reconstruction of instances.
Compared with explicit representations\cite{oleynikova2017voxblox} \cite{vizzo2022sensors} and implicit representation\cite{sitzmann2020implicit}, our approach demonstrates a favorable balance between memory efficiency and high-quality reconstruction. Furthermore, significant opportunities exist for improvement if additional geometric priors are incorporated.
\bibliographystyle{plain}  
\bibliography{ref} 
%


 




\vfill

\end{document}